# Learning Bayesian Network Structure from Correlation-Immune Data

**Eric Lantz**
Computer Sciences Dept.
Biostat. and Med. Informatics Dept.
University of Wisconsin-Madison
Madison, WI 53706

**Soumya Ray**
School of Electrical Engineering
and Computer Science
Oregon State University
Corvallis, OR 97331

**David Page**
Biostat. and Med. Informatics Dept.
Computer Sciences Dept.
University of Wisconsin-Madison
Madison, WI 53706

## Abstract

Searching the complete space of possible Bayesian networks is intractable for problems of interesting size, so Bayesian network structure learning algorithms, such as the commonly used Sparse Candidate algorithm, employ heuristics. However, these heuristics also restrict the types of relationships that can be learned exclusively from data. They are unable to learn relationships that exhibit "correlation-immunity", such as parity. To learn Bayesian networks in the presence of correlation-immune relationships, we extend the Sparse Candidate algorithm with a technique called "skewing". This technique uses the observation that relationships that are correlation-immune under a specific input distribution may not be correlation-immune under another, sufficiently different distribution. We show that by extending Sparse Candidate with this technique we are able to discover relationships between random variables that are approximately correlation-immune, with a significantly lower computational cost than the alternative of considering multiple parents of a node at a time.

## 1 INTRODUCTION

Bayesian networks (BNs) are an elegant representation of dependency relationships present over a set of random variables. The structure of the network defines a factored probability distribution over the variables and allows many inference questions over the variables to be answered efficiently. However, there are a super-exponential number of possible network structures that can be defined over $n$ variables, and the process of finding the optimal structure consistent with a given data set is NP-complete (Chickering et al., 1994), so an exhaustive search to find the one that best matches the data is generally not possible. Techniques to learn BN structure from data must choose a way to restrict the search space of possible networks in order to gain tractability.

The most computationally efficient search technique traditionally employed to discover BN structure is a greedy search over candidate networks. Given a current network, a greedy search scores structures derived using local refinement operators, such as adding and deleting arcs, according to a score such as penalized likelihood. The search keeps only the best such structural modification to refine in the next iteration. It is well understood that greedy approaches are not guaranteed to find global optima. However, another problem with greedy approaches is that of "myopia" in the search. This refers to the fact that such approaches can be confounded when the local changes they consider do not improve the score, even though these changes are relevant to the underlying relationships in the data. In such cases, the search algorithm may not be able to distinguish relevant refinements from irrelevant refinements and may be led astray. Myopic behavior in search can be induced, for example, when the relationship being learned is "correlation-immune". Correlation-immune (CI) functions (Camion et al., 1992) exhibit the property that when all possible function inputs and outputs are listed (for example in a truth table), there is zero correlation between the outputs and all subsets of the inputs of size at most $c$. Examples include exclusive-OR, parity, and consensus, among others. In BN terminology, a child node's probability of taking any particular setting is unchanged when conditioned on at most $c$ of its parents. CI relationships present a problem for greedy approaches to BN structure learning.

The traditional solution to discovering relationships not visible with a greedy search is to consider multiple actions at each step using lookahead (a greedy search could be considered 0-step lookahead). In BNs, looka-head is equivalent to considering multiple changes at once to the parent set of a node. However, lookahead has a very high computational cost that becomes intractable for many interesting problems. In prior work, a technique known as "skewing" has been introduced, that provides many of the benefits of lookahead at a reduced computational cost. This technique relies on the observation that relationships that are CI under a specific input distribution may not be CI under another, sufficiently different distribution. When used with greedy decision tree learners, it was empirically observed that the technique was able to accurately learn functions that were CI under the uniform distribution with only modest amounts of training data (Page & Ray, 2003). Further, the computational cost is at most a linear factor (in the number of variables) over standard greedy tree learning algorithms.

In the present work, we extend the commonly used Sparse Candidate algorithm for BN structure learning to use the skewing technique. We empirically evaluate our algorithm on synthetic data sets generated by different network topologies, both with and without CI relationships. Our results show that, in most cases, our algorithm recovers the generating network topology more accurately than the standard Sparse Candidate algorithm. Further, the networks learned by our approach are significantly more accurate when CI relationships are present in the data.

## 2 BACKGROUND

In this section, we review correlation immunity, the Sparse Candidate (SC) algorithm for learning BN structure and the "skewing" technique.

### 2.1 CORRELATION IMMUNITY

Consider a Boolean function $f$ over $n$ Boolean variables, $x_1, \ldots, x_n$. We say that $f$ is *correlation-immune of order $c$* (Camion et al., 1992; Dawson & Wu, 1997) if $f$ is statistically independent of any subset $S_c$ of variables of size at most $c$: $\Pr(f=1|S_c) = \Pr(f=1)$. An example of such a function is 2-variable odd parity, shown in Figure 1(a). This function is correlation-immune of order one (or equivalently, a first order CI function), because for any subset of variables of size zero or one, the distribution over the values of $f$ does not change. CI functions have been studied extensively in cryptography, where they are used as non-linear ciphers that are not subject to correlation attacks, in which the attacker measures statistics of the output as a method of gaining information about the key. In our work, we are interested in learning approximate CI relationships from data. Figure 1(b) shows a fragment of a BN, where the conditional probability tables

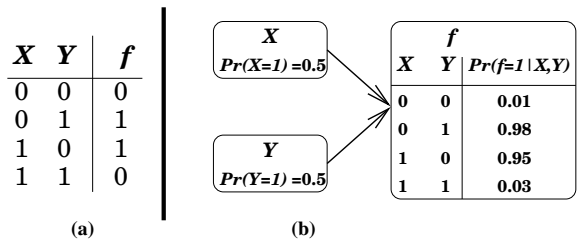

Figure 1: (a) Example of a correlation-immune function. (b) Example of an approximate correlation-immune relationship.

(CPTs) of $X$, $Y$ and $f$ encode an approximate CI relationship between these variables.

CI relationships appear in many real-world scenarios. For example in *Drosophila* (fruit fly), whether the fly survives is known to be an exclusive-OR function of the fly's gender and the expression of *SxL* gene (Cline, 1979). Similarly, during brain development in quail chicks, the *Fgf8* gene, which is responsible for organizing the midbrain, is expressed only in regions where neither or both of the genes *Gbx2* and *Otx2* are expressed (Joyner et al., 2000). This behavior is an instance of *antagonistic repressors* – *Gbx2* and *Otx2* are repressors of *Fgf8*; however, they are also antagonistic – when they are both expressed, they repress each other. Such functions also arise in problems outside of genetics. For example, consider the task of predicting whether two proteins bind to each other. An important predictor of binding is the presence of regions in the proteins that are oppositely charged. Such a function is an exclusive-OR of features representing the charge on regions of the proteins: like charges repel, and thus hinder binding, while opposite charges attract, and thus facilitate binding.

The presence of (approximate) CI relationships in the data presents a challenge for machine learning algorithms that rely on greedy search to gain computational efficiency, such as the SC algorithm described below. This is because at some point in the search, no single feature appears to be relevant to the learning problem. To discover such relationships, *depth-c lookahead* can be used (Norton, 1989). This approach constructs all subsets of $c+1$ features, and will find any target relationship that is correlation-immune of order at most $c$. However, the computational cost is exponential in $c$ ($O(n^{2^{c+1}-1})$ where $n$ is the number of variables), thus this approach can only be used to find small CI functions. Further, this procedure can result in overfitting to the training data, even when only lookahead of depth 1 is considered, because it examines so many alternatives during search (Murthy & Salzberg, 1995; Quinlan & Cameron-Jones, 1995).

## 2.2 THE SPARSE CANDIDATE ALGORITHM

In our work, we are interested in learning probabilistic relationships between the attributes describing the data. To do this, we use the well-known Sparse Candidate algorithm (Friedman et al., 1999), which we review here. The algorithm controls the structure search by limiting the number of parents that will be considered for each variable. The algorithm begins with an initial structure, typically one with no edges. It proceeds by alternating between two phases: a ***restrict*** phase to decide which variables will be considered potential parents (candidates) of each variable, and a ***search*** phase in which greedy structure modifications are made using the candidates and existing structure. The entire algorithm terminates when a search phase fails to make any changes to the structure.

The restrict phase performs a simple test on all pairs of variables in order to reduce the number of actions that need to be considered in the next phase. It limits each variable to a maximum of $k$ candidate parents. For example, if node $Y$ is a candidate parent of node $X$, the next phase of the algorithm will consider adding the directed arc $Y \to X$. The measure of the strength of the correlation between the two variables is the information theoretic measure conditional mutual information $I(X; Y|\mathbf{Z})$ as estimated from the data.

$$I(X; Y|\mathbf{Z}) = \sum_x^X \sum_y^Y \sum_\mathbf{Z}^\mathbf{Z} \hat{p}(x, y, \mathbf{z}) \, log \frac{\hat{p}(x, y|\mathbf{z})}{\hat{p}(x|\mathbf{z})\hat{p}(y|\mathbf{z})} \quad (1)$$

$\mathbf{Z}$ is the set of parents of $X$. If $X$ has no existing parents, $\mathbf{Z} = \phi$ and the equation becomes mutual information. $\hat{p}(x, y|\mathbf{z})$ is the observed joint probability of $x$ and $y$ given the settings of $\mathbf{z}$.

Mutual information (or its conditional variant) is calculated for each pair of variables. For each variable, the current parents are added to the candidate set. Then the candidates with the highest (conditional) mutual information are added until the candidate set the contains $k$ variables. The restrict phase outputs the list of $k$ candidates for each variable.

The search phase consists of a loop to greedily build the best network given the current candidate sets. There are three search operators: add an arc to a variable from one of its candidate parents, remove an existing arc, or reverse the direction of an arc. Each addition or reversal is checked to ensure that directed cycles are not created in the network. All remaining actions are scored, and the best action is taken. Common scoring metrics, including Bayesian-Dirichlet metric (BD) (Heckerman et al., 1995) and Bayesian Information Criterion (BIC) (Schwarz, 1978) include some way of trading off data likelihood with model simplicity. The important criterion of the metric for computational efficiency is that it be decomposable – the contribution of a variable to the score is dependent only on itself and its parents. When an action is considered, the score needs to be recalculated only for the variables whose parents have changed.

The search phase continues until the score is not improved by any available action. If changes have been made to the network during this phase, the algorithm then returns to the restrict phase and chooses new candidate sets based on the current network dependencies. If no changes were made, the algorithm terminates.

The SC algorithm has two greedy components. The restrict phase looks only at pairwise relationships between variables when choosing candidates, and the search phase chooses actions based on their local effect on the score. Both of these are limiting factors in learning approximate correlation immune relationships. For example, if we have data generated by the network fragment in Figure 1(b), the restrict phase of Sparse Candidate is unlikely to select $X$ (or $Y$) as a candidate parent of $f$ (unless there are no other variables in the model) because the mutual information score will be close to zero. Even if $X$ is a candidate parent, the search phase will not add $X$ as a parent of $f$, because doing so will not improve the score of the structure under any of the previously mentioned scoring functions unless $Y$ is already a parent of $f$.

## 2.3 SKEWING

In this section, we review prior work on a technique called "skewing" that has been proposed to learn CI functions in the context of decision tree induction. In the following section, we describe how this approach can be applied to learning structure for BNs.

The motivation for the skewing technique (Page & Ray, 2003) lies in the following observation. Consider a data set over a hundred features, $x_1, \ldots, x_{100}$, where the target function is two variable exclusive-OR, say $x_{99} \oplus x_{100}$. This task is clearly very difficult for a greedy tree learning algorithm. Now, suppose the data are distributed differently from uniform. For example, we might suppose all variables are independent as in the uniform distribution, but every variable has probability only $\frac{1}{4}$ of taking the value 0. In this case, with a large enough sample we expect that the class distribution among examples with $x_{99} = 0$ will differ significantly from the class distribution among examples with $x_{99} = 1$. On the other hand, every variable other than $x_{99}$ or $x_{100}$ is likely to have nearly zero correla-tion with the target. Hence unless a highly unlikely sample is drawn, a greedy tree learning algorithm will choose either $x_{99}$ or $x_{100}$ as the split variable, at which point the remainder of the learning task is trivial.

The desired effect of the skewing procedure is that the skewed data set should exhibit significantly different frequencies from the original data set. To achieve this, the frequency distributions for variables are changed by attaching various weights to the examples as discussed below. In previous work, it was observed that in contrast to skewing, other methods of reweighting (such as boosting or data perturbation (Elidan et al., 2002)) or resampling (such as bagging) did not make CI functions easier to learn (Page & Ray, 2003).

The skewing procedure initializes the weight of every example to 1. For the $j^{th}$ variable, $1 \leq j \leq n$, a "favored setting" $v_j$, either 0 or 1, is selected randomly, uniformly and independently of other variables. The weight $w_i$ of each example in which the $j^{th}$ feature takes the value $v_j$ is changed by multiplying it by a constant $\frac{1}{2} < s < 1$, while the weight of each example in which the feature does not take the value $v_j$ is changed by multiplying it with $1 - s$:

$$w_i = \prod_{j=1}^{n} J(D_{ij}, v_j) \cdot s + (1 - J(D_{ij}, v_j)) \cdot (1 - s), \quad (2)$$

where $D_{ij}$ denotes the $j^{th}$ feature of example $i$ in the data set and $J(x, y)$ is an indicator function that returns one if $x = y$ and zero otherwise. Notice that this reweighting process requires no knowledge of the variables relevant to the target. At the end of this process, it is likely that each variable has a significantly different weighted frequency distribution than previously. But this is not guaranteed, because an unfortunate choice of settings could lead to the new frequency distribution being similar to the old one. A second difficulty is that this process can magnify idiosyncrasies in the original data by assigning some data point with an extremely high weight.

The difficulties in the preceding paragraph occur with some data sets combined with some choices of favored settings. Therefore, instead of using skewing to create only a second distribution, $T$ additional distributions, for small values of $T$, are created. The $T$ different distributions come about from randomly independently selecting $T$ different combinations of favored settings for the $n$ variables according to a uniform distribution.

Each of the $n$ variables is scored for each of the $T + 1$ weightings of the data (the original data set plus $T$ reweighted versions of this data set). The variable that has the largest average score across all skews for the greatest number of weightings is selected as the split variable. The selected variable is highly likely to be *correct* in the sense that it is actually a part of the target function. Yet, in contrast to lookahead, the run-time has been increased only by a constant.

In prior work, it was proven that, given a full truth table of an arbitrary Boolean function, skewing identifies variables relevant to the function with probability 1 (Rosell et al., 2005). Further, empirical evaluation with decision trees has shown that the approach is at least as accurate as standard tree induction algorithms over a large, randomly sampled set of Boolean functions, and shows significantly improved accuracy when the sample is drawn from CI Boolean functions. In our current work, we extend the SC algorithm with this technique to learn BN structure from CI data.

## 3 SKEWING IN SPARSE CANDIDATE

As discussed in section 2.2, the Sparse Candidate algorithm has two greedy steps. The restrict step chooses the $k$ variables most highly correlated with each variable. This step is greedy in the sense that only direct dependence is considered. As we have seen, CI functions show no dependence, so this step will fail to discover relationships. A skewed distribution is created by randomly selecting $\frac{1}{2} < s < 1$ and a preferred setting $v_j$ for all variables. From this, a vector of weights $\vec{w}$ can be calculated for all examples in the training set by equation 2. We can then define the probability of a variable $X$ taking on a certain value $x$.

$$\hat{p}_{skew}(X = x) = \frac{\sum_i w_i | D_{iX} = x}{\sum_i w_i} \quad (3)$$

This reduces to standard frequency counts when all weights are set to 1. We score the correlation between two variables by averaging the skewed conditional mutual information (equation 1) over $T_1 - 1$ skews plus the original distribution, for a total of $T_1$ distributions.

$$I_{skew}(X;Y|\mathbf{Z}) = \frac{\sum_{t}^{T_1} I(X;Y|\mathbf{Z}, \vec{w_t})}{T_1} \quad (4)$$

where $I(X;Y|\mathbf{Z}, \vec{w_t})$ is computed by substituting the $\hat{p}_{skew}$ in equation 3 into $\hat{p}$ in equation 1. Similarly, the search step evaluates each of its possible actions (adding an arc from a variable to one of its candidate parents, removing an arc, or reversing an arc) and chooses the best one according to a decomposable scoring function. Even if a relevant parent is chosen as a candidate in the restrict step, the scoring function – which looks at statistics of the original distribution – will still score the action poorly. So skewing is also needed when evaluating each action. We generate $T_2 - 1$ additional skewed distributions and apply a modified scoring function that takes into account $\vec{w}$. The BD metric calculates the number of times that variables $X$ and $Y$ are found in the training set taking on each combination of their settings. By definition, this equals $\sum_i(1|D_{iX} = x, D_{iY} = y)$, which becomes $\sum_i(w_i|D_{iX} = x, D_{iY} = y)$ with skewing. As in the restrict step, we take the average of the structure scores over all skews before choosing the next action.

The phases of the SC algorithm are shown in Algorithms 1 and 2, with changes due to skewing shown in bold. The first "skewed" distribution in both phases is the original distribution, represented by using a vector of ones for the weight. In both phases, the calculations are affected by the vector $\vec{w}$ produced in creating the skewed distribution. Taking the average result over all skewed distributions serves to preserve the signal from strong relationships, but mitigate the effect of spurious relationships which achieve high scores as the result of a particular skew.

Since we are using multiple distributions, it is not clear how to determine the end condition of the search phase. If we score the modified structure against the original distribution within the search phase (as in normal SC), the search may terminate prematurely because the modification may result in a worse scoring structure if it was part of a CI relationship. Continuing as long as the score improves on the skewed distributions is also problematic, as skewing may cause arcs to be added to the network that are irrelevant to the original distribution. We chose to terminate the search phase when the best move has less than half of the improvement of the first move. This puts bounds on the search and requires strong signals for network modification.

The restrict and search phases alternate, just as they do in normal SC, until the score of the network on the original distribution does not improve with a search phase. Throughout this process, the skewing procedure has used a variety of distributions in order to identify relevant parents. Nevertheless, we want to model the true distribution, not the skewed distributions. Therefore the algorithm closes by running normal SC on the original distribution, but using the structure built from skewing as the initial structure. This step could have the effect of removing extra arcs, or it could serve to find the remaining parents of a CI relationship. Our experiments (data not shown) illustrate that this step greatly improves precision by removing unnecessary arcs that poorly model the original distribution.

---

**Algorithm 1**: Sparse Candidate Restrict Phase Adapted from Figure 2 in Friedman et al. (1999). Changes due to skewing shown in bold.

Input: A matrix $D$ of $m$ data points over $n$ variables, number of candidates $k$, initial network $B_\tau$
Output: For each variable $x_i$ a set of candidate parents $c_i$ of size $k$

1　$\vec{w_1} \leftarrow \vec{1}$
2　**for $t \leftarrow 2$ to $T_1$ do　$\vec{w_t} \leftarrow$ Skew(D)**
3　for $i \leftarrow 1$ to $n$ do
4　　**for $t \leftarrow 1$ to $T_1$ do**
5　　　Calculate $I(x_i, x_j|Pa(x_i), \vec{w_t})$ for all $x_j \neq x_i$ and $x_j \notin Pa(x_i)$
6　　end
7　　Choose the $k - l$ variables with the highest $I_{skew}$ **over all skews**, where $l = |Pa(x_i)|$
8　　Set $c_i = Pa(x_i) \cup \{k - l$ chosen variables $\}$
9　end
10　return $\{c_i\}$

---

**Algorithm 2**: Sparse Candidate Search Phase Changes due to skewing shown in bold.

Input: A matrix $D$ of $m$ data points over $n$ variables, initial network $B_\tau$, candidate parents $\{c_i\}$
Output: Network $B_{\tau+1}$

1　$\vec{w_1} \leftarrow \vec{1}$
2　**for $t \leftarrow 2$ to $T_2$ do　$\vec{w_t} \leftarrow$ Skew(D)**
3　repeat
4　　$B_{\tau+1} \leftarrow B_\tau$
5　　**for $t \leftarrow 1$ to $T_2$ do**
6　　　Calculate $Score(B_\tau, action|D, \vec{w_t})$ for all possible actions
7　　end
8　　Apply action with highest **average score over all skews** to $B_\tau$
9　until *Score improvement threshold not met (see text)*
10　return $B_{\tau+1}$

---

Since the CPT of a node representing an exact CI function is simply a truth table, we can apply the theory from previous work (Rosell et al., 2005) that skewing is always able to identify a relevant variable (variable involved in the CI relationship) if given complete data. With only a sample of the data the outcome is no longer certain, but has previously been shown to occur consistently.

It is difficult to compute computational complexity of the SC algorithm or its skewed variant, due to the unknown number of iterations. However, we can say something about the effect of skewing on the complexity of each phase. The restrict phase of SC is $O(n^2)$, where $n$ is the number of variables, due to the calculation of pairwise (conditional) mutual information scores. With skewing, it becomes $O(T_1 n^2)$. The search phase will undergo an unknown number of iterations, but the process of choosing the action is $O(kn)$. Skewing raises that to $O(T_2(kn))$. Thus the effect of skewing on the computational complexity is linear in the number of skews used in each phase.

## 4 EXPERIMENTS

In this section, we discuss the evaluation of the effectiveness of skewing in the context of two types of graph structures, with or without CI relationships. We expect skewing to have a strong advantage over normal SC when CI relationships are present in the generating network, and that advantage will also be present with approximate CI relationships. Additionally, we expect skewing to not decrease the effectiveness of SC in networks which do not contain CI relationships.

For all experiments, we constructed Bayesian networks of Boolean variables. Training data and test data were sampled uniformly from the network. We set $T_1 = 30$, $T_2 = 30$, $k = 6$, and the test sets contained 1000 samples. The skewing weight factor, $s$, was randomly chosen in each skew. To account for the randomness implicit in the algorithm, skewed SC was run 5 times on each network. The scoring metric used in the search was K2 (Cooper & Herskovits, 1992), a version the BD metric, with a structure term that penalized based on the number of parameters in the network and the size of the training set ($\sum_i 2^{|pa(i)|} log|D|/2$).

We used two measures to evaluate the effectiveness of our algorithm on synthetic data. The first is the log likelihood of the model given the test data, which describes how well the data appears to have been generated by the model. We also wanted to look at whether the correct arcs of the generating structure were being discovered by the algorithm. Unfortunately, most CI functions are statistically invariant as to which variable is the "output". For example, Figure 1(a) could represent $f = X \oplus Y$, $X = Y \oplus f$, or $Y = f \oplus X$, and the difference is impossible to determine solely from data. So instead of looking for the exact directed arcs, we compare the Markov blankets of the generating structure and learned structure. The Markov blanket of a variable X consists of X's parents, X's children, and the other parents of X's children. The Markov blankets for all variables will be the same in all output variations of CI functions. In order to penalize both missing and superfluous arcs, we calculate the $F1$ score of the Markov blanket of all variables. Precision is the fraction of Markov blanket variables returned by the algorithm that are present in the generating structure. Recall is the fraction of Markov blanket variables in the generating structure that are returned by the algorithm.

$$F1 = \frac{2 \cdot precision \cdot recall}{precision + recall} \qquad (5)$$

The first synthetic network type consisted of 30 variables, with one variable having 5 parents and all others having no parents. For all parents P(X=1) = 0.5, whereas the probabilities of other variables were randomly assigned. The CPT of the child variable represented either a CI function or a function constructed by randomly selecting the output for each row of the truth table. The function representation could be exact (probability of 1 for the function value and 0 otherwise) or approximate (function value having a probability of 0.9 or 0.8).

Figure 2 shows learning curves for these experiments as a function of the number of examples in the training set. In terms of both likelihood and Markov blanket F1 score, skewing greatly outperforms normal SC on CI data sets. The difference between the two algorithms on the exact functions is statistically significant at the 99.9% confidence level by a two-tailed t-test under both measures when the training set size $\geq$ 400. Normal SC fails to improve despite more training data being available. Interestingly, skewed SC also outperforms normal SC for randomly generated functions. This can be explained by noting that randomly generated functions could contain CI subproblems (of 2, 3, or 4 variables) or be CI functions themselves. The difference between the two algorithms on the exact functions is significant at 95% confidence for training set size $\geq$ 1600. Skewing shows some robustness to approximate CI relationships, particularly when measured by Markov blanket F1 score. However, for every 10% reduction in the probability of the CPT returning the function value, scores fall by more than half as compared to the baseline in all cases.

Another synthetic network type was inspired by the Quick Medical Reference (QMR) network structure (Shwe et al., 1991) as a representation of disease diagnosis. The structure consists of a layered bipartite graph, with directed arcs only from the top layer to the bottom layer. The bottom layer nodes represent symptoms. The nodes in the top layer represent diseases or conditions, with arcs towards the symptoms they influence. It is possible to imagine conditional probability tables for the nodes in the lower level that are CI functions like exclusive-OR or 3 variable parity. The generating networks contained 20 top layer variables and 20 bottom layer variables, with bottom layer nodes having 2 or 3 parents. We examined how well normal and skewed SC could reconstruct the structures with varying probability that a given bottom layer node would have a CPT that represented a CI function. Figure 3(a) shows that skewing outperforms normal SC as measured by Markov blanket F1 score, and while both algorithms suffer as more CI CPTs are present in the generating structure, skewing continues to be more accurate. When all bottom layer nodes have CI function CPTs, normal SC is unable to discover any true arcs.

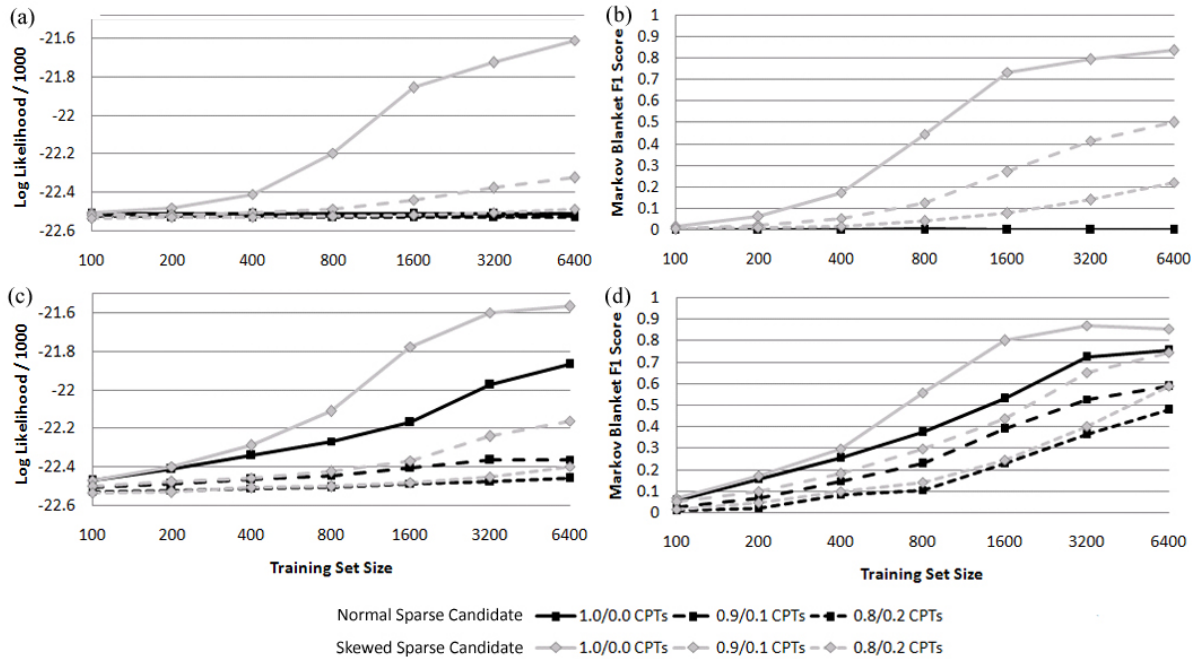

Figure 2: Learning curves on 30-variable data sets. Each data point is the average of 100 generated data sets. (**a-b**) Performance of normal and skewed SC on CI 5 variable functions as measured by log likelihood (**a**) and Markov blanket F1 (**b**). (**c-d**) Performance on random 5 variable functions by log likelihood (**c**) and Markov blanket F1 (**d**).

Additionally, we considered the effect of adding prior knowledge to the structure learner in the form of labeling the nodes as belonging in the top or bottom layer, and allowing arcs only from top layer to the bottom layer. Since the algorithms are prevented from making certain types of errors, we would expect this to improve scores. Figure 3(b) shows that the Markov blanket F1 scores are indeed improved for both versions of SC, but the performance of skewing now improves as more of the nodes are CI functions, and the Markov blanket F1 (which is very close to the F1 of the returned structure due to the limitations on allowed arcs) reaches 0.975. In both graphs, skewed SC outperforms normal SC even when there are no CI relationships present.

## 5   CONCLUSIONS

The commonly used SC algorithm employs greedy heuristics to learn BN structure. While efficient, the search is myopic and can be led astray if variables describing the data have CI relationships. We have presented an approach that integrates skewing with the SC algorithm. Our experiments demonstrate that this extension enables SC to learn accurate network structures from CI data, at a lower computational cost than considering multiple parents for each variable during the search. The skewing approach can outperform normal SC on different graph types and in cases where there are no CI relationships.

In current work, we are using our approach to learn regulatory networks from gene expression data sets. As we noted in Section 2.1, examples of CI relationships often appear in genetics. We expect that our approach will be useful in learning relationships between genes in such cases. However, analyzing gene expression data is difficult for several reasons. First, the data tends to be very sparse and high-dimensional, since the expression levels of thousands of genes are measured for each experiment and there are typically few experiments. High-dimensional sparse data tends to be problematic for the skewing approach, because the reweighting process tends to magnify idiosyncrasies of the sample. Further, gene-expression data tends to be noisy. In our experiments, we observed that our approach can handle only small amounts of noise. Making our approach more robust is an important direction for future work. Finally, gene-expression data, like many other real-world data sets, has variables that are continuous-valued. We can only apply our current approach to this data by discretizing it. Prior work has investigated approaches that use the skewing technique with continuous and nominal variables. Inte-grating such approaches into the BN structure learning framework is an important direction for future work.

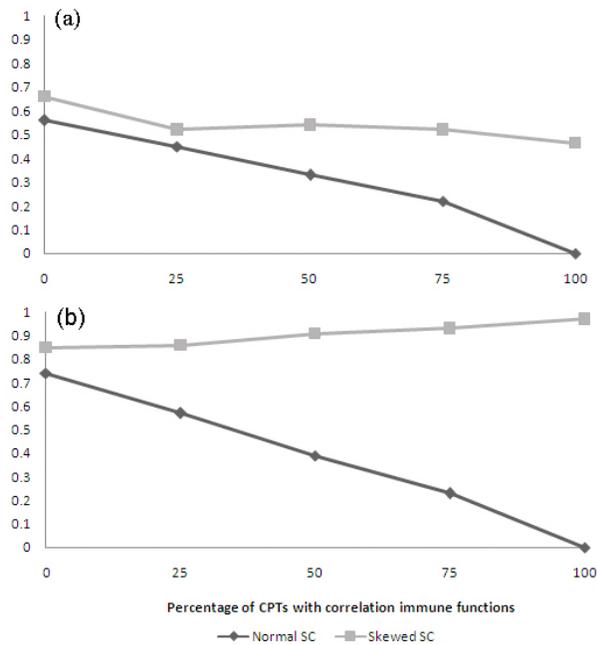

Figure 3: Markov blanket F1 scores on synthetic QMR-like data as function of percentage of CPTs with CI functions. (**a**) Unconstrained learning. (**b**) Constrained to only allow arcs from top layer to bottom layer.

**Acknowledgments**

This research was supported by NSF IIS0534908. EL was also supported by NLM 5T15LM007359. Thanks to Eric Bach for his help with correlation immunity.